# Extracting Synonyms from Bilingual Dictionaries

**Mustafa Jarrar**
Birzeit University
Palestine
mjarrar@birzeit.edu

**Eman Karajah**
Birzeit University
Palestine
1105486@student.birzeit.edu

**Muhammad Khalifa**
Cairo University
Egypt
m.khalifa@grad.fci-cu.edu.eg

**Khaled Shaalan**
The British University in Dubai
United Arab Emirates
khaled.shaalan@buid.ac.ae

## Abstract

We present our progress in developing a novel algorithm to extract synonyms from bilingual dictionaries. Identification and usage of synonyms play a significant role in improving the performance of information access applications. The idea is to construct a translation graph from translation pairs, then to extract and consolidate cyclic paths to form bilingual sets of synonyms. The initial evaluation of this algorithm illustrates promising results in extracting Arabic-English bilingual synonyms. In the evaluation, we first converted the synsets in the Arabic WordNet into translation pairs (i.e., losing word-sense memberships). Next, we applied our algorithm to rebuild these synsets. We compared the original and extracted synsets obtaining an F-Measure of 82.3% and 82.1% for Arabic and English synsets extraction, respectively.

## 1 Introduction

The importance of synonyms is growing in a number of application areas such as computational linguistics, information retrieval, question answering, and machine translation among others. Synonyms are also considered essential parts in several types of lexical resources, such as thesauri, wordnets (Miller et al., 1990), and linguistic ontologies (Jarrar, 2021; Jarrar, 2006).

There are different notions of synonymy in the literature varying from strict to lenient. In ontology engineering (see e.g., Jarrar, 2021), synonymy is a formal equivalence relation (i.e., reflexive, symmetric, and transitive). Two terms are synonyms *iff* they have the exact same concept (i.e., refer, intentionally, to the same set of instances). Thus, $T_1 =_{Ci} T_2$. In other words, given two terms $T_1$ and $T_2$ lexicalizing concepts $C_1$ and $C_2$, respectively, then $T_1$ and $T_2$ are considered to be synonyms *iff* $C_1 = C_2$. A less strict definition of synonymy is used for constructing Wordnets, which is based on the substitutionablity of words in a sentence. According to Miller et al. (1990), “two expressions are synonymous in a linguistic context *c* if the substitution of one for the other in *c* does not alter the truth value”. Others might refer to synonymy to be a “closely-related” relationship between words, as used in distributional semantics, or the so-called *word embeddings* (see e.g., Emerson, 2020). Word embeddings are vectors of words automatically extracted from large corpora by exploiting the property that words with a similar meaning tend to occur in similar contexts. But it is unclear what type of similarity word vectors capture. For example, words like red, black and color might appear in the same vector which might be misleading synonyms.

Extracting synonyms automatically is known to be a difficult task, and the accuracy of the extracted synonyms is also difficult to evaluate (Wu et al., 2003). In fact, this difficulty is also faced when modeling synonyms manually. For example, words like *room* (غرفة) and *hall* (قاعة) are synonyms only in some domains like schools and events organization (Daher et al., 2010). Indeed, synonymy can be general or domain-specific; and since domains and contexts are difficult to define (Jarrar, 2005), and since they themselves may overlap, constructing a thesaurus needs special attention.

Another difficulty in the automatic extraction of synonyms is the polysemy of words. A word may have multiple meanings, and its synonymy relations with other words depend on which meaning(s) the words share. Assume $e_1$, $e_2$, and $e_3$ are English words, and $a_1$, $a_2$ and $a_3$ are Arabic words, we may have $e_1$ participating in different synonymy and translation sets, such as, $\{e_1, e_2\}=\{a_2\}$ and $\{e_1, e_3\}=\{a_3\}$. For example {table, tabular array}={جدول} and {river, stream}={جدول}.

In this paper, we present a novel algorithm to automatically extract synonyms from a given bilingual dictionary. The algorithm consists of two phases. First, we build a translation graph and extract all paths that form cycles; such that, all nodes in a cycle are candidate synonyms. Second, cyclic paths are consolidated, for refining and improving the accuracy of the results. To evaluate this algorithm, we conducted an experiment using the Arabic WordNet (AWN) (Elkateb et al., 2016). More specifically, we built a flat bilingual dictionary, as pairs of Arabic-English translations from AWN. Then, we used this bilingual dictionary as input to our algorithm to see how much of AWN's synsets we can rebuild.

Although the algorithm is language-independent and can be reused to extract synonyms from any bilingual dictionary, we plan to use it for enriching the Arabic Ontology - an Arabic wordnet with ontologically clean content (Jarrar, 2021; Jarrar, 2011). The idea is to extract synonyms, and thus synsets, from our large lexicographic database which contains about 150 Arabic-multilingual lexicons (Jarrar at el, 2019). This database is available through a public lexicographic search engine (Alhafi et al., 2019), and represented using the W3C lemon model (Jarrar at al., 2019b).

This paper is structured as follows: Section 2 overviews related work, Section 3 presents the algorithm, and Section 4 presents its evaluation. Finally, Section 5 outlines our future directions.

## 2 Related work

Synonyms extraction was investigated in the literature mainly for constructing new Wordnets or within the task of discovering new translation pairs. In addition, as overviewed in this section, some researchers also explored synonymy graphs for enhancing existing lexicons and thesauri.

A wordnet, in general, is a graph where nodes are called *synsets* and edges are *semantic relations* between these synsets (Miller et al., 1990). Each synset is a set of one, or more synonyms, which refers to a shared meaning (i.e., signifies a concept). Semantic relations like hyponymy and meronymy are defined between synsets. After developing the Princeton WordNet (PWN), hundreds of Wordnets have been developed for many languages and with different coverage (see *globalwordnet.org*).

Many researchers proposed to construct Wordnets automatically using the available linguistic resources such as dictionaries, wiktionaries, machine translation, corpora, or using other Wordnets. For example, Oliveira and Gomes (2014) proposed to build a Portuguese Wordnet automatically by building a synonymy graph from existing monolingual synonyms dictionaries.

Candidate synsets are identified first; then, a fuzzy clustering algorithm is used to estimate the probability of each word pair being in the same synset. Other approaches proposed to also construct wordnets and lexical ontologies via cross-language matching, see (Abu Helou et al., 2014; Abu Helou et al., 2016).

A recent approach to expand wordnets, by Ercan and Haziyev (2019), suggests to construct a multilingual translation graph from multiple Wiktionaries, and then link this graph with existing Wordnets in order to induce new synsets, i.e., expanding existing Wordnets with other languages.

Other researchers suggest to use dictionaries and corpora together, such as Wu and Zhou (2003) who proposed to extract synonyms from both monolingual dictionaries and bilingual corpora. First, a graph of words is constructed if a word appears in the definition of the other, and then assigned a similarity rank. Second, a bilingual English-Chinese corpus (pairs of translated sentences) is used to find links between words if they appear in the same pair, with a probability rank. Third, a monolingual Chinese corpus is used to find words co-occurring in the same context. These three results are then combined together using the ensemble method. A more recent approach, by Khodak et al. (2017), proposed to use an unsupervised method for automated construction of Wordnets using PWN, machine translations, and word embeddings. A target word is first translated into English using machine translation, and these translations are used to build a set of candidate synsets from PWN. Each candidate synset is then ranked with a similarity score that is calculated using the word embedding-based method.

A similar attempt to build Arabic and Vietnamese Wordnets was proposed by Lam et al. (2014). They proposed a method to automatically construct a new Wordnet using machine translation and existing Wordnets. Given a synset in one or more Wordnets, all words in this synset (in multiple languages) are translated using machine translation into the target language. The retrieved translations, which contain wrong translations because of polysemy, are ranked based on their relative frequencies, and the highest ranked translations are retrieved. This approach was extended by Al-Tarouti et al. (2016) by introducing word embeddings to better validate and remove irrelevant words in synsets.

Other related work to synonymy extraction is the task of finding new translations, such that, given translation pairs between multiple languages, one may discover new translation pairs that are not explicitly stated. For example, Villegas et al. (2016) presented an experiment to produce new translations from a translation graph constructed from the Apertium dictionaries. Given a set of multilingual translation pairs, a translation graph is constructed, from which cycles are extracted. New translation pairs are then identified if they participate in the extracted cycles. The experiment illustrated that some wrong translations might be detected because of polysomy, thus a path density score was assigned to each path, such that low densities are excluded. More recently, Torregrosa et al. (2019) presented three algorithms for automatic discovery of translations from existing dictionaries, namely, cycle-based, path-based, and multi-way neural machine translation. In the cycle-based approach, a translation graph is constructed from a multilingual dictionary, and cycles of length 4 are identified. However, in the path-based approach, a frequency weight is assigned to each path based on the number of translation pairs participating in this path, such that paths of lower length and higher frequency get lower weights. In the third algorithm, multilingual parallel corpora were used to train a multi-way neural machine translation, and continued the training based on the output of the other two algorithms. An experiment by the authors shows a very low recall and a reasonable precision (25-75%) for the three approaches.

The main differences between these approaches and our approach, is that we aim at extracting synonyms rather than translation pairs, and that we assume the translation graph to be formed of nodes from two languages only. Having two languages in the translation graph produces a different number of paths; thus, different disambiguation complexity.

A lexicon-based algorithm called CQC was proposed by Flati and Navigli (2012). The algorithm takes a bilingual dictionary as input, then builds a translation graph, from which only cyclic and quasi-cyclic paths are extracted. These paths are then ranked, such that shorter paths are given higher ranks than longer ones. Words, and words senses, encountered in the cycles or quasi-cycles are likely to be synonymy candidates. This approach is mainly used for validating and enriching the Ragazzini-Biagi English-Italian dictionary, but it can be also used for extracting synonyms. The accuracy of this approach depends on the structure of input dictionaries, which is assumed to contain senses, e.g., an English word and its set of equivalent Italian translations. This implies that these Italian words are themselves synonyms.

In our approach, we assume that a word in a given language is translated to only one word in the other language, i.e., only translation pairs, without synonymy relations. In other words, we assume the bilingual input to be the most ambiguous.

As will be discussed in Section 4, our algorithm does not assume any pre-existing conditions or assumptions about the input data, and does not use part-of-speech or any other morphological features. Designing an algorithm without any pre-existing assumption, makes the algorithm more reusable (Jarrar et al., 2002) for other languages and other types of lexicons. Nevertheless, and as described in the future work Section, using linguistic features would improve the algorithm's accuracy.

## 3 Our Algorithm

The problem we aim to tackle in this paper is described as the following: given a set $\boldsymbol{B}$ of bilingual translation pairs of the form ($a_i$, $e_j$), where $a_i$ is a word in language $l_1$ and $e_j$ is its translation in language $l_2$. Our goal is to extract a set $\boldsymbol{R}$ of bilingual synonyms, such that $\{a_1,..,a_k\} = \{e_1,..,e_l\} \in \boldsymbol{R}$.

To extract the set $\boldsymbol{R}$ of bilingual synonyms from $\boldsymbol{B}$, our algorithm performs two steps:

***Step 1: Extract cyclic paths***

Given $\boldsymbol{B}$, an undirected graph is built, where each node represents a word of either language and two edges (in both directions) connect any two nodes that represent a word-translation pair. Then, we use Johnson's algorithm (Johnson, 1977) to find all cycles in the directed graph. A cycle is a path of nodes that starts and ends in the same node, such as $a_1 \rightarrow e_1 \rightarrow a_2 \rightarrow e_2 \rightarrow a_1$. Nodes participating in the same path are considered candidate synonyms, and converted into bilingual synsets, e.g., $\{a_1, a_2\} = \{e_1, e_2\}$. To avoid very long cycles, we modify Johnson's algorithm to stop expanding a path beyond the pre-specified maximum cycle length $k$. Figure 1 illustrates an Arabic-English translation graph extracted from the Arabic WordNet. The graph starts from the word ġābaẗ (غابَة), expands its English translations, then expand the Arabic translations of each English word, and so on, up to 7 levels ($k$=7).
The expansion stops in these cases:

1) The root node is found, i.e., cycle,
2) No more translations are found, which are underlined (e.g., woodland), or
3) The max $k$ level is reached.

The output of this step is a set of candidate bilingual synsets extracted from the nodes participating in cyclic paths, such as:

1. {forest, woods} = {غابَة, غاب}
2. {forest, woods} = {غابَة, أدْغال}
3. {forest, wood} = {غابَة, أدْغال}
4. {forest, wood}={غابَة, غاب}
5. {wood, woods } = {غابَة, غاب}
6. {wood, woods} = {غابَة, أدْغال}
7. {forest, wood, woods} = {أدْغال, غابَة, غاب}

***Step 2: Consolidation***

This step aims to merge synsets that have the same sets of translations. In other words, Arabic synsets are consolidated (i.e., unioned) if they have the same English synsets, such as synsets 1 and 2, 3 and 4, and 5 and 6, in the previous example. Similarly, English synsets are consolidated if they have the same Arabic synsets. This step is repeated until no more consolidations are found. The output of this phase is the final sets of bilingual synonyms, such as:

{forest, wood, woods} $=_{ci}$ {أدْغال, غابَة, غاب}

As will be shown in the evaluation section, the consolidation phase is important, in order to minimize the impact of paths that might not be sufficiently expanded, if $k$ is small. The consolidation phase is designed based on the following heuristics:

*(i)* It is less likely for a set of bilingual synsets, especially long synsets, to refer to multiple concepts. In other words, the longer a synset (i.e., more bilingual words in the synset), the less likely the synset to be polysemous and to refer to multiple meanings.

*(ii)* It is less likely that a synset, especially long synsets, to be a subset of another synset. That is, it is possible in practice to have different synsets, like $\{a, b, c, d\}$ and $\{a, b, c\}$, where the former is a subset of the later, which may negatively affect the accuracy of our algorithm. However, such cases are less likely to happen, especially in case of long synsets.

*(iii)* It is less likely for the same English synset, especially long synsets, to be translated into multiple Arabic synsets. An English synset may have multiple concepts and thus multiple Arabic synsets, such as $\{e_1, e_2, e_3, e_4\}=_{c1}\{a_1, a_2, a_3, a_4\}$ and $\{e_1, e_2, e_3, e_4\}=_{c2}\{a_5, a_6, a_7, a_8\}$. Such cases may negatively affect our accuracy, but they are rare in practice.

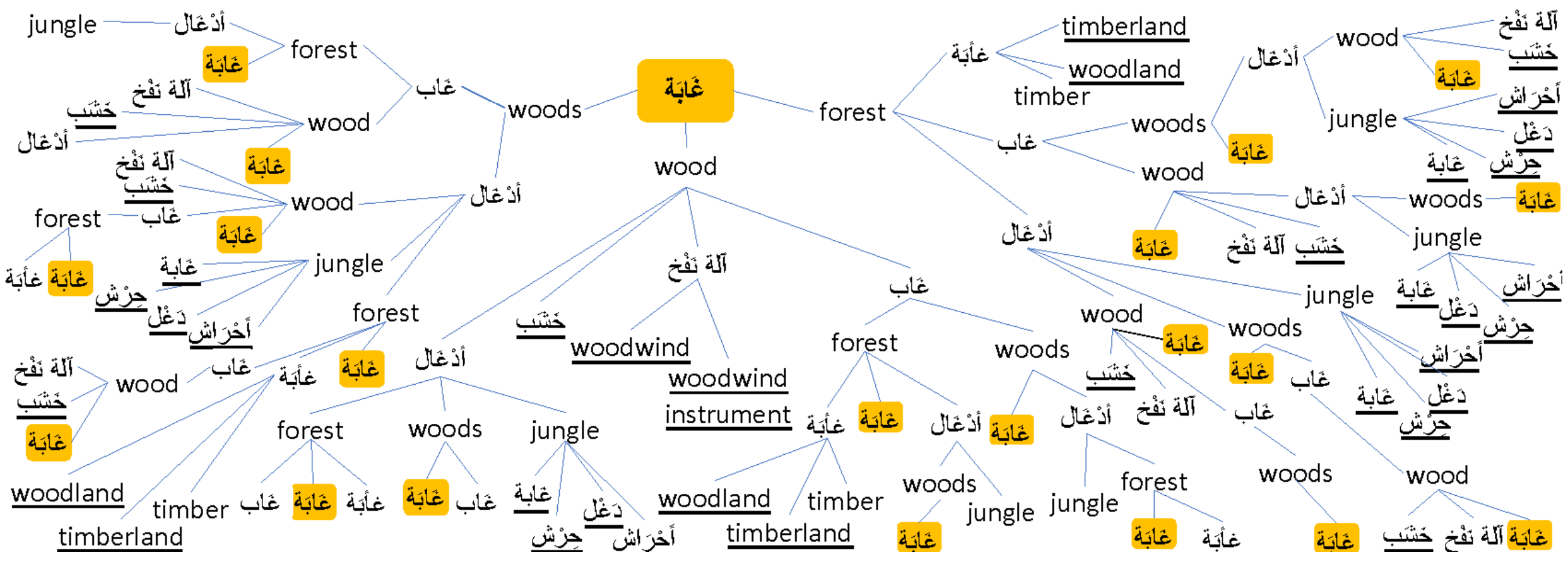

**Figure 1: Example of a translation graph**

## 4 Evaluation and Discussion

To evaluate the extent to which our synonyms extraction algorithm is able to produce correct results, we used the Arabic Wordnet (AWN). The AWN, which is a set of bilingual synsets, is converted into a flat bilingual dictionary (i.e., translation pairs), such that all synonymy links between words are lost. Then, we use our algorithm to restore these links (extract bilingual synsets), and compare the extracted with original synsets.

Our choice of evaluating the algorithm using the AWN data is because this resource contains highly polysemous words, and thus evaluates the algorithm in challenging cases. In addition, and in order to evaluate the algorithm in unguided conditions or any pre-existing assumption, and make it reusable for other languages, we did not apply any fine-tuning or language-specific preprocessing or treatment. Thus, we assume that the input translation pairs do not have any tag indicating their part of speech (POS) or other morphological features, or whether words are MSA or dialect (Jarrar et al., 2017). We also assume that Arabic words with different diacritic signs, even if they are compatible (Jarrar et al., 2018), are different words. For example (غَايَة) and (غايَة) are considered different words because of slightly different diacritics. Tuning the algorithm to take into account such morphological features, inflections, and diacritics, would very likely improve the accuracy of the results; but this is not a goal in this paper and is left as a future work.

As evaluation metrics, we use the precision, recall and F-measure to compare the extracted synsets with the original AWN as the gold standard. We use the Cosine similarity to compute the match between two given synsets.

For precision, we count the number of correctly extracted synsets divided by all the extracted synsets. In cases of partial match between two synsets $x$ and $y$, we use the max similarity with all the gold sets as the "correctness" of the extracted synset:

$$Precision = \frac{\sum_{x \in extracted} max_{y \in AWN} Cosine(x,y)}{|Extracted\ synsets|}$$

where $Cosine(x,y) \in [0,1]$. Recall and F-measure are computed as:

$$Recall = \frac{\sum_{y \in AWN} max_{x \in Extracted} Cosine(x,y)}{|AWN|}$$

$$F\text{-}Measure = 2 * \frac{Precision\ .Recall}{Precision + Recall}$$

Tables 1 and 2 show the evaluation metrics when extracting Arabic and English synsets, respectively. Clearly, the proposed consolidation step has a positive effect on the algorithm by boosting the F-measure from 74.4% to 82.3% and from 70.1% to 82.1% for Arabic and English, respectively for a path length *k*=6.

| | Precision | Recall | F-Measure |
|---|---|---|---|
| k=6, no consolidation | 62.5 | **91.9** | 74.4 |
| k=6, with consolidation | **80.5** | 84.2 | **82.3** |
| k=8, with consolidation | 64.4 | 84.3 | 73.0 |

**Table 1: Results on the AWN for Arabic synsets extractions using the proposed algorithm.**

| | Precision | Recall | F-Measure |
|---|---|---|---|
| k=6, No consolidation | 57.6 | **89.8** | 70.1 |
| k=6, with consolidation | **80.4** | 83.8 | **82.1** |
| k=8, with consolidation | 64.7 | **84.0** | 73.1 |

**Table 2: Results on the AWN for English synsets extractions using the proposed algorithm**

The results also show that having longer paths (e.g., *k*=8) does not improve the accuracy, which is most likely in case of highly polysemous words, where some irrelevant nodes are generated in longer paths.

Last but not least, the Arabic Wordnet contains about 10K synsets, and most of the words in these synsets are, by definition, highly polysemous Arabic and English words. This is because these 10K synsets are called Common Base concepts, and assumed to be frequently used and exist in many languages. As discussed earlier such high polysemy is likely to affect the accuracy; thus, evaluating our algorithm on less polysemous words is likely to produce better accuracy.

## 5 Conclusions and Future Work

We presented our progress in developing a novel algorithm to extract synonyms from bilingual dictionaries. Although the algorithm was evaluated on extracting English-Arabic bilingual synsets, it is reusable for other languages, especially since it does not assume any language-specific treatment or preprocessing. Our choice of using AWN in the evaluation, which contains highly polysemous words, illustrates that our algorithm produces realistic results in such challenging cases.

We plan to extend our algorithm in different directions. We plan to take into account part of speech tags and other morphological features when generating candidate synonyms. Similarly, words with different, but compatible, diacritics, inflections, and forms need a special treatment. Such extensions and fine-tunings are very likely to produce higher accuracy.

## Acknowledgments

This research is partially supported by the Research Committee at Birzeit University.

## References

Alhafi, D., Deik, D., & Jarrar, M. (2019): Usability Evaluation of Lexicographic e-Services. In Proceedings – 2019 IEEE/ACS 16th International Conference on Computer Systems and Applications, Abu Dhabi (pp.1-7). IEEE. doi:10.1109/AICCSA47632.2019.9035226

Daher, J., & Jarrar, M. (2010). Towards a Methodology for Building Ontologies – Classify by Properties. In Proceedings – 3rd Palestinian International Conference on Computer and Information Technology (PICCIT), Palestine.

Elkateb, S., Black, W., Vossen, P., Farwell, D., Pease A., & Fellbaum, C. (2006). Arabic WordNet and the Challenges of Arabic. In Proceedings – Arabic NLP/MT Conference (pp. 665-670).

Emerson, G. (2020). What are the Goals of Distributional Semantics. In Proceedings of the 58th Annual Meeting of the Association for Computational Linguistics. ACL. (pp. 7436-7453).

Ercan, G., & Haziyev, F. (2019). Synset expansion on translation graph for automatic wordnet construction. Information Processing & Management, 56(1), 130-150.

Helou, M. A., Palmonari, M., & Jarrar, M. (2016). Effectiveness of Automatic Translations for Cross-Lingual Ontology Mapping. Journal of Artificial Intelligence Research, 55, 165-208. doi:10.1613/jair.4789

Helou, M. A., Palmonari, M., & Jarrar, M., Fellbaum, F. (2014). Towards Building Lexical Ontology via Cross-Language Matching. In Proceedings – 7th Conference on Global WordNet. Global WordNet Association. (pp. 346–354). EID: 2-s2.0-84859707947

Jarrar, M., & Meersman, R. (2002). Scalability and Knowledge Reusability in Ontology Modeling. In Proceedings – International Conference on Advances in Infrastructure for Electronic Business, Science, and Education on the Internet (SSGRR 2002s). Scuola Superiore G Reiss Romoli. Rome, Italy.

Jarrar, M. (2005). Towards Methodological Principles for Ontology Engineering. Doctoral dissertation, Vrije Universiteit Brussel, Belgium.

Jarrar, M., (2006). Towards the Notion of Gloss, and the Adoption of Linguistic Resources in Formal Ontology Engineering. In Proceedings – 15th international conference on World Wide Web, (pp.497-503). ACM. doi: 10.1145/1135777.1135850

Jarrar, M. (2011): Building A Formal Arabic Ontology (Invited Paper). In Proceedings – Experts Meeting on Arabic Ontologies and Semantic Networks, Tunis. ALECSO, Arab League.

Jarrar, M., Habash, N., Alrimawi, F., Akra, D., & Zalmout, N. (2016). Curras: An Annotated Corpus for the Palestinian Arabic Dialect. Language Resources and Evaluation, 50(219), 1-31. doi:10.1007/S10579-016-9370-7

Jarrar, M., Zaraket, F., Asia, R., & Amayreh, H. (2018). Diacritic-based Matching of Arabic Words. ACM Transactions on Asian and Low-Resource Language Information Processing (TALLIP), 18(2), 1-21. doi: 10.1145/3242177

Jarrar, M., & Amayreh, H. (2019). An Arabic-Multilingual Database with a Lexicographic Search Engine. In Proceedings – 24th International Conference on Applications of Natural Language to Information Systems (NLDB 2019). Lecture Notes in Computer Science (vol. 11608, pp. 234-246). Springer. Doi:10.1007/978-3-030-23281-8_19

Jarrar, M., Amayreh, H., & McCrae, J. (2019): Representing Arabic Lexicons in Lemon – a Preliminary Study. In Proceedings – 2nd Conference on Language, Data and Knowledge, Leipzig, Germany. CEUR-WS (vol. 2402, pp. 29-33).

Jarrar, M. (2021). The Arabic Ontology - An Arabic Wordnet with Ontologically Clean Content. Applied Ontology Journal, IOS Press.

Johnson, D. B. (1975). Finding all the elementary circuits of a directed graph. SIAM Journal on Computing, 4(1), 77-84.

Khodak, M., Risteski, A., Fellbaum, C., & Arora, S. (2017). Automated WordNet construction using word embeddings. In Proceedings of the 1st Workshop on Sense, Concept and Entity Representations and their Applications (pp. 12-23).

Lam, K., Tarouti, F., & Kalita J. (2014). Automatically constructing Wordnet synsets. In Proceedings of the 52nd Annual Meeting of the Association for Computational Linguistics (Volume 2: Short Papers) (pp. 106-111).

Miller, J., Beckwith, R., Fellbaum, C., Gross D., & Miller, K. (1990). Introduction to Wordnet: An on-line Lexical Database. International Journal of Lexicography, 3(4), 235-244.

Oliveira, H., & Gomes, P. (2014). ECO and Onto. PT: a flexible approach for creating a Portuguese wordnet automatically. Language resources and evaluation, 48(2), 373-393.

Tarouti, F., & Kalita, J. (2016). Enhancing automatic wordnet construction using word embeddings. In Proceedings – Workshop on Multilingual and Cross-lingual Methods in NLP (pp. 30-34).

Tiziano, F., & Navigli. R. (2012). The CQC algorithm: Cycling in graphs to semantically enrich and enhance a bilingual dictionary. Journal of Artificial Intelligence Research, 43, 135-171.

Torregrosa, D., Mihael, A., Ahmadi, S., & McCrae, J. (2019). TIAD 2019 Shared Task: Leveraging knowledge graphs with neural machine translation for automatic multilingual dictionary generation. Translation Inference Across Dictionaries.

Villegas, M., Melero, M., Gracia J., & Bel, N. (2016). Leveraging RDF graphs for crossing multiple bilingual dictionaries. In Proceedings of the Tenth International Conference on Language Resources and Evaluation (LREC'16) (pp. 868-876).

Wu, H., & Zhow M. (2003). Optimizing synonym extraction using monolingual and bilingual resources. In Proceedings of the second international workshop on Paraphrasing (pp. 72-79).